\documentclass{article}

\usepackage{PRIMEarxiv}

\usepackage[utf8]{inputenc} % allow utf-8 input
\usepackage[T1]{fontenc}    % use 8-bit T1 fonts
\usepackage{hyperref}       % hyperlinks
\usepackage{url}            % simple URL typesetting
\usepackage{booktabs}       % professional-quality tables
\usepackage{amsfonts}       % blackboard math symbols
\usepackage{nicefrac}       % compact symbols for 1/2, etc.
\usepackage{microtype}      % microtypography
\usepackage{lipsum}
\usepackage{fancyhdr}       % header
\usepackage{graphicx}       % graphics
\usepackage{algorithm}
\usepackage{algorithmic}
\usepackage{amsmath}
\graphicspath{{media/}}     % organize your images and other figures under media/ folder

 \usepackage{multirow}

\title{EEE, Remediating the failure of machine learning models via a network-based optimization patch
}

\author{
   Ruiyuan Kang \\
  Department of Mechanical Engineering \\
  Khalifa University \\
  Abu Dhabi, UAE\\
  \texttt{ruiyuan.kang@ku.ac.ae} \\
       \And
  Dimitrios C. Kyritsis \\
  Department of Mechanical Engineering \\
  Research and Innovation Center on CO$_2$ and Hydrogen \\
  Khalifa University \\
   Abu Dhabi, UAE\\
  \texttt{dimitrios.kyritsis@ku.ac.ae} \\
   \And
  Panos Liatsis* \\
  Department of Electrical Engineering and Computer Science \\
  Khalifa University \\
   Abu Dhabi, UAE\\
  \texttt{panos.liatsis@ku.ac.ae} \\
}

\begin{document}
\maketitle

\begin{abstract}
A network-based optimization approach, EEE, is proposed for the purpose of providing validation-viable state estimations to remediate the failure of pretrained models. To improve optimization efficiency and convergence, the most important metrics in the context of this research, we follow a three-faceted approach based on the error from the validation process. Firstly, we improve the information content of the error by designing a validation module to acquire high-dimensional error information. Next, we reduce the uncertainty of error transfer by employing an ensemble of error estimators, which only learn implicit errors, and use Constrained Ensemble Exploration to collect high-value data. Finally, the effectiveness of error utilization is improved by using ensemble search to determine the most prosperous state. The benefits of the proposed framework are demonstrated on four real-world engineering problems with diverse state dimensions. It is shown that EEE is either as competitive or outperforms popular optimization methods, in terms of efficiency and convergence.
\end{abstract}

% keywords can be removed
\section{Introduction}
A Machine Learning (ML) model $f_p$ is often trained to map from observations $\mathbf{y} \in \mathcal{Y}$ to states $\mathbf{x}\in \mathcal{X}$, i.e., $f_p: \mathcal{Y} \to \mathcal{X}$. However, $f_p$ cannot be assured to be thoroughly reliable as its performance is predominantly determined by the information in the training data, and moreover, the model parameters are fixed after training. Making mistakes is fine for general-purpose applications, however, in most science and engineering problems, providing unreliable solutions may be unacceptable. Examples of such instances may be when $f_p$ provides physically unreasonable estimates in airplane design \cite{kou2021data}, or predicts zero carbon emissions when processing the information in the flame of hydrocarbon combustion \cite{hanEnsembleDeepLearning2022}.

A variety of approaches attempted to improve the reliability of $f_p$, such as adversarial attacking \cite{baiRecentAdvancesAdversarial2021}, special network structure design \cite{zhangRethinkingLipschitzNeural},  mega data/model \cite{floridiGPT3ItsNature2020}, symbolic learning \cite{cranmerDiscoveringSymbolicModels2020},  domain-knowledge regularization \cite{raissiPhysicsinformedNeuralNetworks2019}, learning neural operator \cite{liFourierNeuralOperator2021}, etc. The common aspect of the aforementioned methods is that they attempt to improve the quality of learning, and subsequently, increase the trustworthiness of  $f_p$. In this work, a different angle is pursued, i.e., a means of assessing the quality of the mapping through the introduction of a validation module $f_v$, which assesses whether $\hat{\mathbf{x}}$ is acceptable via feedback $e \in \mathcal{E}$ (hereafter, the error), i.e., $f_v: \mathcal{X} \to \mathcal{E}$. This could be thought as being analogous to confirming student responses to questions during an exam. In Natural Language Processing (NLP), validation is frequently performed through training verifier \cite{liAdvanceMakingLanguage2022}, human judgment and guidance (prompt) \cite{weiChainThoughtPrompting2022}, and recently, training with self-verification \cite{wengLargeLanguageModels2022}. In science and engineering applications, trustworthy validations are often provided by rules, laws, and physical models that are proven theoretically or experimentally.

In fact, a more critical issue is how to obtain a validation-viable $\hat{\mathbf{x}}$ to address the potential failure of $f_p$. A natural way forward is to develop a trustworthy Complementary System (CS) so as to assure reliability, although this may not provide the same levels of efficiency and convenience as $f_p$. In the works of ~\cite{murturiDECENTDecentralizedConfigurator2022,chenRuntimeSafetyAssurance2022}, the CS was respectively realized by a physical model and a predefined solution database. Notably, the use of an optimization method may be an option ~\cite{kangSelfValidatedPhysicsEmbeddingNetwork2022} in cases with a tolerance of iterations, which is the focus of this research. In a typical scenario, a pre-collected dataset may not be available, as $f_v$ often changes and depends on a given observation $\mathbf{y}_0$. This is commonly seen in design problems \cite{degiorgiJetEngineDegradation2019,hollermannFlexibleHereandnowDecisions2021}, and thus obtaining $e$ from $f_v$ via queries is the method of choice. Considering the fact that $f_v$ usually involves time-consuming physical simulation processes, the requirement for an ideal CS can be described as providing a validation-viable $\hat{\mathbf{x}}$ with the highest possible success rate and the smallest possible number of queries to the validation module, i.e., convergence and efficiency.

\begin{figure*}[hbt!]
\centering
\includegraphics[width=1.0\textwidth]{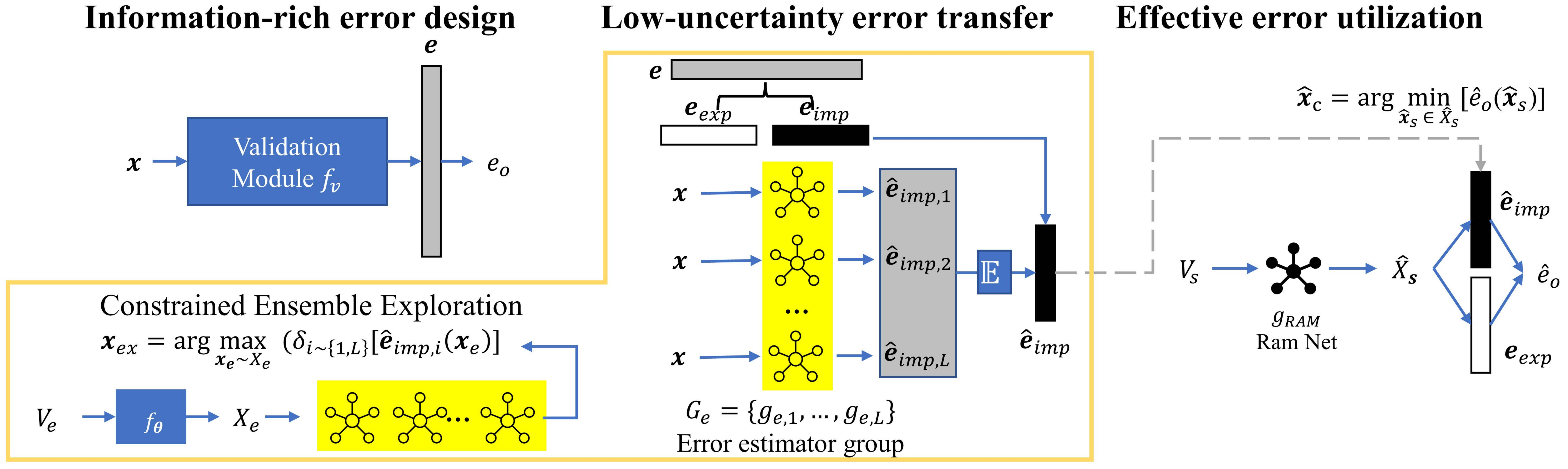}
\caption{Three facets of error utilization in EEE}
\label{fig1}
\end{figure*}

This work proposes a CS module based on a network-based optimization method. The method is named Ensemble search and Ensemble exploration with an Ensemble of estimators (EEE). The core principle is broadly based on a combination of the "predict then optimize" approach \cite{mandiSmartPredictandOptimizeHard2019,fannjiangAutofocusedOraclesModelbased}, and online active exploration \cite{shyamModelBasedActiveExploration2019}, however, with special considerations on the requirements of efficiency and convergence. The means to meet these two requirements is the error from $f_v$, as it provides the bridge between $f_v$ and CS. Correspondingly, the main contribution of this paper is to provide an alternative interpretation and use of the error, as follows: (1) Construction of information-rich error functional via the specialist design of $f_v$; (2) transfer of high certainty error information via error categorization, an ensemble of error estimators, and constrained ensemble exploration; (3) Effective and efficient error utilization via ensemble state search and appropriate selection of state search model type. Systematic experimental studies were performed in four problems with diverse state dimensions. The results demonstrate that EEE is as competitive as or outperforms state-of-the-art optimization methods, while being applicable to problems of varying dimensionality. An abstract representation of the proposed contributions is presented in Fig.\ref{fig1}, with more details provided in Sec.\ref{EEE}.

\section{Related Work}
This work attempts to tackle the important issue of reliability of pretrained machine learning (PML) models $f_p$. An online optimization system is needed to provide a validation-viable solution as soon as $f_p$ fails. Therefore, instead of pursuing an optimal solution, in this work, we attempt to provide acceptable solutions, however, meeting the critical requirements of high efficiency and convergence. 

\textbf{Model-augmented Optimization} One popular way to utilize network models for solving optimization problems is based on a pretrained network-based proxy model of validation module or inverse mapping of $f_p$, which maps states to errors or observations, i.e., $\mathcal{X} \to \mathcal{E}$ or $\mathcal{Y}$, using  optimization algorithms, such as Particle Swarm Optimization (PSO) \cite{vaezinejadHybridArtificialNeural2019,han2022surrogate}, Genetic Algorithm (GA) \cite{hegdeAcceleratingOpticsDesign2019} to be the search model that iterates the input of the proxy model in order to search for an acceptable $\hat{\mathbf{x}}$ for a given $\mathbf{y}_0$. In addition to a hybrid architecture, some works directly use another network or the original $f_p$ to be the search model \cite{liuTrainingDeepNeural2018,brookesConditioningAdaptiveSampling,chenNeuralOptimizationMachine2022,villarrubiaArtificialNeuralNetworks2018,peurifoyNanophotonicParticleSimulation2018a}. However, since the proxy model is a PML, its reliability/authority cannot be guaranteed, which is exactly the reason why CS is needed.

\textbf{Offline Model-based Optimization} Offline model-based optimization (MBO) is a topic of increasing interest that hypothesizes that a pre-collected dataset $B$ exists, which contains the information of $\mathcal{X} \to \mathcal{E}$. The task is to infer the potentially optimal $\hat{\mathbf{x}}$, which can lead to minimum error. A popular two-step paradigm is "predict then optimize" \cite{elmachtoubSmartPredictThen2020,fannjiangAutofocusedOraclesModelbased,mandiSmartPredictandOptimizeHard2019}, where a proxy model is trained on $B$ to simulate $f_v$, and a search model is simultaneously trained to find the optimal $\hat{\mathbf{x}}$. One potential problem with this approach is that it may overfit the information in $B$. To avoid this, adversarial samples, noise-added weight, and inexact estimations are used \cite{yuRoMARobustModel, fuOfflineModelBasedOptimization2021,trabuccoConservativeObjectiveModels}. Apart from this approach, Model Inversion Network (MIN) \cite{kumarModelInversionNetworks2019} treats the optimization problem as a generative problem from error to state, i.e., $\mathcal{E} \to \mathcal{X}$, which requires MIN has reliable extrapolation strength because the minimum value of $e$ is often outside $B$. Recently, the development of Global Optimization Network (GON) \cite{zhaoGlobalOptimizationNetworks} appears to solve the extrapolation issue by embedding dedicated unimodal functions into a lattice neural network. However, a massive amount of data is needed to fix the lattice neural network. In the context of the current research, the key issue is that there is lack of a pre-collected set $B$, and thus, convergence and efficiency are the main priorities rather than pursuing the optimal solution.

\textbf{Online Model-based Optimization} Most model-based Reinforcement Learning (RL) algorithms \cite{haarnojaSoftActorCriticAlgorithms2018} aim at solving optimization problems in an online fashion. RL mainly focuses on acquiring an intelligent policy model, however, in our approach, the objective is acquiring validation-viable states $\hat{\mathbf{x}}$ in an efficient and stable manner. Ma et al.\cite{maDeepFeedbackInverse2020} utilized a pretrained optimizer model to guide the correction of state estimation. However, if $f_p$ is lacking in reliability, it is not possible to obtain such an authoritative optimizer. Some works attempted to make $f_v$ differentiable so that $e$ can be directly obtained without the need of a proxy model \cite{chandrasekharTOuNNTopologyOptimization2021,amosOptNetDifferentiableOptimization,raissiPhysicsinformedNeuralNetworks2019}. However, in real-world science and engineering 
 problems, it is often difficult or even impossible for the models in $f_v$ to be fully differentiable. Although MIN \cite{kumarModelInversionNetworks2019} proposed active data acquirement through Thompson sampling \cite{russo2018tutorial}, this reduces to Bayesian optimization \cite{pmlr-v180-daulton22a} with neural network priors \cite{snoek2015scalable}, and cannot outperform Bayesian optimization with non-parametric Gaussian process\cite{williams2006gaussian} in terms of efficiency \cite{kumarModelInversionNetworks2019}. A conceptually similar work is SVPEN \cite{kangSelfValidatedPhysicsEmbeddingNetwork2022}, which treats the solution of inverse problems in engineering applications as the optimization of forward problems via neural networks. However, it does not appear to consider convergence and efficiency.

\section{EEE: Ensemble Search
and Ensemble Exploration with an Ensemble of Estimators}
\label{EEE}
Prior to describing EEE, the general symbol representation rules are provided as follows: ordinary lowercase letters, such as $e$, represent  scalars, functions and networks; bold lowercase letters, such as $\mathbf{e}$, represent vectors or tensors; capital letters, such as $G$, represent sets, groups or subspaces; and calligraphic letters, such as $\mathcal{X}$, represent spaces.

EEE consists of four modules: (1) the validation module $f_v$ is used for assessing the quality of $\hat{\mathbf{x}}$, (2) the error estimators $G_e$ are used to estimate the error function, (3) the state exploration module is used to collect state-error data to support the training of $G_e$, and (4) the RAM net $g_{RAM}$ is the search model, used to search validation-viable states. Notably, the final selected state $\hat{\mathbf{x}}$ for a given observation $\mathbf{y}_0$ has to go through $f_v$ to assure reliability.

As aforementioned, efficiency and convergence are the key metrics that distinguish EEE from existing network-related optimization methods. From our perspective, the means to satisfy  these requirements is the error function. In RL or offline MBO, the error/reward/objective is often a scalar $e$, and can only be passively acquired from a black-box environment. This error is then transferred by a network-based proxy model to the search model. This approach makes the error ambiguous and affected by uncertainty, thus harming the efficiency and convergence of state search. In contrast, the validation module $f_v$ in EEE is specifically designed to assure the reliability of $\hat{\mathbf{x}}$. Therefore, the structure of $f_v$ is at least partially transparent, which differs from the conventional concept of the environment. In EEE, the core consideration is utilizing the information contained in the error function to improve the efficiency and convergence of state search. More specifically, let us consider that $f_v$ produces the error function with low ambiguity, $G_e$ transfers this error with low uncertainty, and $g_{RAM}$ uses it with high effectiveness. The pseudocode of EEE in given in Algorithm \ref{alg1}, and its specific processes are described in the following sections.

% \begin{wrapfigure}{L}{0.45\textwidth}
% \begin{minipage}{0.45\textwidth}
\begin{algorithm}
\caption{Overview of EEE framework}
\label{alg1}
\begin{algorithmic}[1]
\REQUIRE{Validation module $f_v$, Accuracy threshold $\epsilon$}, Maximum iteration round $r_{max}$
\ENSURE{acceptable state $\hat{\mathbf{x}}$: $e_o(\hat{\mathbf{x}}) < \epsilon$}
\STATE initialize RAM net $g_{RAM}$, state seeds $V_s$, error estimator group $G_e$, exploration buffer $B_e$
 \FOR{Iteration $r<r_{max}$}
    \FOR [\hfill\COMMENT{\textbf{step 1} Train error estimators}] {$i<r_{ep}$} 
        \STATE train $g_e \in G_e$ on $B_{sub} \subset B_e\cup B_h$ 
        \STATE stop train $g_e$ \textit{if} $\mathcal{L}_{g_e,i}< \epsilon_{es}$
    \ENDFOR
    \STATE count the number of convergent models $l$
    \STATE $r_{RAM} \propto l$ \hfill\COMMENT{\textbf{step 2} Search state}
    \FOR {Round $j<r_{RAM}$}
        \STATE $\hat{e}_o(\mathbf{x})=\mathbf{w}_{exp}^T\mathbf{e}_{exp}+\mathbf{w}_{imp}^T\mathbb{E}_{g_e\sim G_e}[g_e(\mathbf{x})]$
        \STATE  update $g_{RAM}$ via min $\mathcal{L}_{RAM}={\mathbb{E}_{\mathbf{v}_s \sim V_s}(\hat{e}_o(g_{RAM}(\mathbf{v}_s)))}$
    \ENDFOR
    \STATE $\hat{\mathbf{x}}_c=arg \underset{\hat{\mathbf{x}} \sim \hat{X}_s}{\min}[\hat{e}_o(\hat{\mathbf{x}})]$
    \IF [\hfill\COMMENT{\textbf{step 3} Validate state}]{$\hat{e}_o(\hat{\mathbf{x}}_c)<c_f*\epsilon$}
     \STATE $e_o(\hat{\mathbf{x}}_c),e_{imp}(\hat{\mathbf{x}}_c) \gets f_v(\hat{\mathbf{x}}_c)$ 
     \STATE historical buffer $B_e \gets \hat{\mathbf{x}}_c,e_{imp}(\hat{\mathbf{x}}_c), e_o(\hat{\mathbf{x}}_c)$
     \STATE $\hat{\mathbf{x}}=\hat{\mathbf{x}}_c$ \textit{if} $e_o(\hat{\mathbf{x}}_c <\epsilon)$, stop
    \ENDIF
    \STATE $X_e=f_{\theta}(V_e)$ \hfill\COMMENT{\textbf{step 4} Explore state}
     \STATE $\mathbf{x}_{ex}= arg max_{\mathbf{x}_e \in X_e}\left\{\sigma_{g_e \sim G_e} [g_e(\mathbf{x}_e)]\right\}$
    \STATE $e_o(\mathbf{x}_{ex}),e_{imp}(\mathbf{x}_{ex}) \gets f_v(\mathbf{x}_{ex})$
    \STATE Exploration buffer $B_e \gets \mathbf{x}_{ex},e_{imp}(\mathbf{x}_{ex}), e_o(\mathbf{x}_{ex})$
\ENDFOR

\end{algorithmic}
\end{algorithm}
% \end{minipage}
% \end{wrapfigure}
\subsection{Validation Module Design}

The purpose of the validation module $f_v$ is two-fold. First, it plays the role of referee, by assessing the quality of $\hat{\mathbf{x}}$. Second, it is also the guide, since the information of the error from $f_v$ should have the highest level of fidelity so as to guide state search. 

Both of these roles require $f_v$ to provide a high-dimensional error vector $\mathbf{e}$, rather than merely a scalar $e$. In regards to the role of referee, it is natural that validation should be done from different levels of the hierarchy (e.g., at system and subsystem) with various criteria (e.g., weight and volume for an object) to tightly constrain the feasible solution space $X_{y_0} \subset \mathcal{X}$, thus assuring that $\mathbf{x}\in X_{y_0}$ is acceptable. Turning to the role of guide, the high dimensionality of the error function translates to higher levels of information content, which can alleviate ill-conditioning, and reduce the number of trials of the state search, since the contribution of the error elements is more informative than a scalar $e$. Typically, when a scalar $e$ exists, regularization terms are added to augment single $e$ into $\mathbf{e}$, and the final $e$ is a weighted projection of $\mathbf{e}$, which correspondingly increases the amount of information in the error function.

The preferred condition is that  the error space $\mathcal{E} \subseteq\mathbb{R}^{D_e}$ is bijective to $\mathcal{X}$. However, increasing the dimensionality of the error function $D_e$ blindly is not always preferred. Under the condition described in Proposition 1, error component $\mathbf{e}_j$ merely provides a finer or equivalent (bijection case) description than $\mathbf{e}_i$, although its suitability to the problem at hand may vary. An intuitive case is the use of \textit{MSE} and \textit{RMSE}.

\textbf{Proposition 1.} For any two error components, $\mathbf{e}_i,\mathbf{e}_j$, if the same partial entries of state $\mathbf{x}_p \subseteq \mathbf{x}$ are successfully validated, then the existence of $\mathbf{e}_i$ is redundant, when $\mathbf{e}_j$ is surjective to $\mathbf{e}_i$, i.e., $f(\mathbf{e}_j(\mathbf{x}_p))=\mathbf{e}_i(\mathbf{x}_p)$.

However, a problem that arises is that the incorporation of an increased dimensionality $\mathbf{e}$ in the validation module design conflicts with the requirement of finding an acceptable $\hat{\mathbf{x}}$. This is due to the intrinsic multi-objective properties of $\mathbf{e}$, which require that optimization methods determine a set of $\hat{\mathbf{x}}$, near the Pareto front. However, determining a single $\hat{\mathbf{x}}$ requires single-objective optimization. Thus, the multi-dimensional components of vector $\mathbf{e}$ can be merged together into an overall error function $e_o$ (i.e., penalty operation \cite{sonmez2011multi}), which, in turn, results to loss of error information. To resolve such a conflict, the use of a neural network is an appropriate option. Its architectural flexibility in terms of multiple outputs can accommodate that $g_e$ approximates multiple components of $\mathbf{e}$. Moreover, a scalar loss function can be defined to represent $e_o$, and set as the loss function for the search model $g_{RAM}$ so as to find a single $\hat{\mathbf{x}}$. Thus, the information content of the error can be maintained, when backpropagating this loss function.

Accordingly, we need to produce a means of constructing the overall error $e_o \in R$, and require that the validation module $f_v$ computes this, in addition to $\mathbf{e}$. A convenient means of doing so is to calculate the weighted sum of the components of $\mathbf{e}$, as shown in Eq.\ref{eq1}.  We set the condition that a potential $\hat{\mathbf{x}}$ can be accepted, if $e_o(\hat{\mathbf{x}}) \le\epsilon$, where $\epsilon$ is a predefined threshold. Consequently, the distribution of the overall error should be either a unimodal function or a multimodal function, however, all states satisfying the aforementioned condition are acceptable.

\begin{equation}
\label{eq1}
    e_o(\mathbf{x})=\sum w_i\mathbf{e}_i(\mathbf{x}_{p,i})
\end{equation}

\subsection{Low-uncertainty Error Transfer}
\label{error transfer}
Although it is possible to transfer all components of $\mathbf{e}$ through $g_e$, the nature of neural networks determines that the approximated error contains a degree of uncertainty, i.e., $\hat{\mathbf{e}}_i=\mathbf{e}_i+\delta_i$, with a standard deviation $\sigma_{\delta, i}$. Suppose, for convenience of analysis, that all $\sigma_{\delta,i}$s have similar amplitude, thus, the overall uncertainty can be represented by $D_e*\sigma_{\delta}$. Consequently, the benefits brought about by the higher information content of the error function may be offset by the larger degree of approximation uncertainty. Thus, a question that naturally arises is how to transfer $\mathbf{e}$ with the smallest possible degree of uncertainty.

\textbf{Implicit and Explicit Errors} The first action is to simplify the problem of error estimation. Indeed, the components of $\mathbf{e}$ are split into two categories according to whether $\mathbf{e}_i(\mathbf{x}_p)$ can be directly back propagated to $\mathbf{x}_p$. This group of error components are termed as explicit errors $\mathbf{e}_{exp}$, while the remaining components are called implicit errors $\mathbf{e}_{imp}$. Thus, Eq.~\ref{eq1} can be expressed as Eq.~\ref{eq2}. Typical choices for explicit errors $\mathbf{e}_{exp}$ are regularization terms or simple differentiable, physics-based, validation operations on state $\mathbf{x}$, so that these operations can be expressed in a differentiable form with little computational expense, In contrast, implicit errors $\mathbf{e}_{imp}$ are often the products of complex, simulation-based, validation processes, and are inherently non-differentiable with respect to state $\mathbf{x}$ or instead,  formulation of a differentiable model is very computationally expensive.

\begin{equation}
\label{eq2}
    e_{o}=\mathbf{w}_{exp}^T\mathbf{e}_{exp}+\mathbf{w}_{imp}^T\mathbf{e}_{imp}
\end{equation}

Thus, $g_e$ needs to only approximate the components of $\mathbf{e}_{imp}$, while $\mathbf{e}_{exp}$ can be directly calculated from $\mathbf{x}$ through differentiation. Naturally, the use of exact signals to replace the simulated signals leads to a reduction in error transfer uncertainty. The concept of error blurriness (EB) is proposed to measure error transfer uncertainty (Eq.~\ref{eq3}). This quantity is defined as the product of the amplitude of $\sigma_{\delta}^2$ and the partition of $\mathbf{e}_{imp}$, and the  smaller its value the better. The ideal case is EB$=0$ where all error components are differentiable, similar to  \cite{chandrasekharTOuNNTopologyOptimization2021,amosOptNetDifferentiableOptimization}, however, this is difficult to realize in practice.
\begin{equation}
\label{eq3}
    EB=\frac{n\sum^n_{j=1} \mathbf{w}_{imp,j}}{m\sum^m_{i=1} \mathbf{w}_{exp,i}+n\sum^n_{j=1} \mathbf{w}_{imp,j}}\sigma_{\delta}^2
\end{equation}

\textbf{Ensemble of Error Estimators} To further reduce error transfer uncertainty, we employ a number, $L$, of error estimators $g_e$s, rather than a single module to construct the set $G_e$, and train them independently using bootstrapped sampling \cite{mooney1993bootstrapping}, using the data collected from the online buffer ($B_e$ and $B_h$, which will be explained later). The sampling set difference and random initialization help each $g_e$ learn differently. Next, their average estimate is used to approximate $\mathbf{e}_{imp}$, as shown in Eq.\ref{eq4}: 
\begin{equation}
\label{eq4}
   \hat{\mathbf{e}}_{imp}=\mathbb{E}_{g_e\sim G_e}(g_e)(\mathbf{x}))
\end{equation}

Therefore, for each element inside $\mathbf{e}_{imp}$,  the corresponding $\sigma_{\delta}^2$ reduces by $1/L$. Hence, error blurriness can be calculated by:
\begin{equation}
\label{eq5}
    EB=\frac{\frac{n}{L}\sum^n_{j=1} \mathbf{w}_{imp,j}}{m\sum^m_{i=1} \mathbf{w}_{exp,i}+\frac{n}{L}\sum^n_{j=1} \mathbf{w}_{imp,j}}\sigma_{\delta}^2
\end{equation}
The functionality of $g_{\mathbf{e}}$ can be realized by minimizing:
\begin{equation}
  \label{eq6}  
  \mathcal{L}_{g_e}=\mathbb{E}_{\mathbf{x}\sim X}[d( g_e(\mathbf{x}),\mathbf{e}_{imp}(\mathbf{x}))]
\end{equation}
where $d$ is a suitable distance function. 
It should be noted that it is possible to approximate specific components $\mathbf{e}_{imp,i}$ by using specific $\mathbf{x}_{p,i}$, if $\mathbf{e}_{imp,i}$ only depends on $\mathbf{x}_{p,i}$.

\textbf{Constrained Ensemble Exploration} The last factor, which can be optimized in EB is the amplitude of $\sigma_{\delta}^2$. This variance term depends on three factors, i.e., the learning capability of $g_e$, the dataset used to train $g_e$, and the choice of training method. In turn, the learning capacity simply translates to $g_e$ having a sufficient number of learnable parameters. The second factor requires effective exploration of $\mathcal{X} \to \mathcal{E}$ to collect the exploration data buffer $B_e$. Last, the choice of learning algorithm will be discussed in detail in Sec. \ref{dynamic scheduel}.

Benefitting from the existence of $G_e$. a state $\mathbf{x}_{ex}= arg max_{\mathbf{x} \sim \mathcal{X}}\left\{\sigma_{g_e \sim G_e} [g_e(\mathbf{x})]\right\}$ can be selected for exploration, and subsequently passed on to $f_v$ to acquire $\mathbf{e}$ for training data collection. The mechanism here is that a state that $G_e$ fails to make consistent estimations is the state that $G_e$ needs to enhance. This state is often selected by training an active exploration network $g_{ex}$ \cite{pathakSelfSupervisedExplorationDisagreement,shyamModelBasedActiveExploration2019}. Nevertheless, the adversarial learning style between $g_{ex}$ and $G_e$ causes $G_e$ to frequently shift focus, and even collapse by chasing challenging states. This shift in focus in $G_e$ also affects the optimization stability of the search model $g_{RAM}$, thus leading to reductions in optimization efficiency and convergence. 

In this work, Constrained Ensemble Exploration (CEE) is proposed to balance  exploration capability and optimization stability. First, the definition of a qualified constrained space generator is given. A function $f_{\theta}: \mathbb{R}^{D_{ex}} \to X_{sub}$ is a qualified constrained space generator if the union of $X_{sub}$ generated by $\theta \sim \Theta$ is $\mathcal{X}$, i.e.,  $\mathcal{X}=\bigcup_{\theta \sim \Theta} X_{sub}(\theta)$.

The principle of CEE is to split $\mathcal{X}$ into sub-spaces $X_{sub}$ by $f_{\theta}$, controlled by pattern parameters $\theta$. Rather than training an active model to search globally, a very large number of candidate states $X_e$ can be sampled by feeding a random number $\mathbf{v}_e \in R^{D_{ex}}$ to $f_{\theta}$, and choosing $\mathbf{x}_{ex}= arg max_{\mathbf{x}_e \in X_e \subset X_{sub}}\left\{\sigma_{g_e \sim G_e} [g_e(\mathbf{x})]\right\}$  to be explored. In this way, the active global exploration is transformed into passive sampling within constrained subspaces, which maintains optimization stability, and waives the requirement for training of the state exploration model. A potential choice of $f_{\theta}$ is a linear function $f_{\theta}:\mathbb{R} \to X_{sub}$, which can be simply approximated by utilizing a single-layer perceptron with re-initialization but no training, so that the exploration is performed on a hyper-line in each optimization round.

\subsection{Efficient Error Utilization}
\textbf{Ensemble State Search} Rather than shifting the focus of the original model $f_p$ \cite{fannjiangAutofocusedOraclesModelbased,kangSelfValidatedPhysicsEmbeddingNetwork2022}, a disposable network, named RAM net,  $g_{RAM}:\mathbb{R}^{D_s} \to \mathcal{X}$,  is used to search the state from the random vector $\mathbf{v}_s \in \mathbb{R}^{D_s}$ without processing $\mathbf{y}_0$. This can be done  because $\mathbf{y}_0$ is unchanged for a given optimization problem and can thus be treated as a prior. Given the decoupling between a given observation and its corresponding state, an ensemble of candidate states $\hat{X}_s$ can be acquired from $g_{RAM}$ by feeding a set of vectors $V_s$: 
 \begin{equation}
   \label{eq14}
     \hat{X}_s=g_{RAM}(V_s|y_0)=g_{RAM}(V_s)
 \end{equation}
 
 The state search is performed by updating $g_{RAM}$ to minimize $\hat{e}_o$, given by:
\begin{equation}
\label{eq13}\hat{e}_o(\mathbf{x})=\mathbf{w}_{exp}^T\mathbf{e}_{exp}+\mathbf{w}_{imp}^T\mathbb{E}_{g_e\sim G_e}[g_e(\mathbf{x})]
\end{equation}
 by using the loss function of Eq.\ref{eq15}: 
 \begin{equation}
\label{eq15}
    \mathcal{L}_{g_{RAM}}=\mathbb{E}_{\mathbf{v}_s \sim V_s}[\hat{e}_o(g_{RAM}(\mathbf{v}_s)]
\end{equation}
 It is notable that $\hat{e}_o$ only contains the error transfer uncertainty from the estimation of implicit error. By using ensemble estimation and direct calculation of $\mathbf{e}_{exp}$, the uncertainty is significantly reduced. Moreover, the entire expression shown in Eq.\ref{eq13}, rather than merely the scalar value of $\hat{e}_o$, is used to guide the state search. Thus, the contribution of each element of $e_o$ is clearly reflected in $g_{RAM}$, thus improving search efficiency. It is worth noting that since there is a set of states $\hat{X}_s$, the scenario where some states are caught in local optima is not of concern.
 
Due to efficiency considerations, querying the entire set of $\hat{X_s}$ is unaffordable. Therefore, we utilize the idea of greedy search, i.e., only the estimated state $\hat{\mathbf{x}}_c$ associated to the minimum $\hat{e}_o$ is fed to $f_v$ for evaluation, as shown in Eq.\ref{eq16}. The validated data is stored in the historical buffer $B_h$, and used together with $B_e$ in training $G_e$.
\begin{equation}
\label{eq16}
    \hat{\mathbf{x}}_c=arg \underset{\hat{\mathbf{x}} \sim \hat{X}_s}{\min}[\hat{e}_o(\hat{\mathbf{x}})]
\end{equation}

\textbf{RAM Net as a Kernel} An alternative view of $g_{RAM}$ is to be regarded as a special type of a kernel function $k_{\theta}$. However, it is not mandatory to use a learnable kernel $k_{\theta}$. Instead, a fixed kernel, such as the identity kernel, may be used, and consequently, since $\mathbf{v}_s \leftrightarrow \hat{x}_s$, the optimization is directly performed on the $\mathbf{v}_s$. In this case, the combination of $g_{RAM}$ and $G_e$ is in fact a special case of an unconditional energy model \cite{lecunEnergyBasedModelsDocument2007}. The use of a learnable kernel, i.e., a network architecture, brings about two important benefits. First, it offers architectural flexibility, whereby the network can incorporate complex structures and functions, and can thus be used to embody information about the nature of the problem. For instance, a ConvNet is more suitable than a Multi-Layer perceptron (MLP) for problems, where the state has sparse local patterns. Second, it offers advantages in terms of learning and memory capabilities. Specifically, the large number of weights in $k_{\theta}$ can be used to learn and store knowledge on manipulating state construction, and can thus realize complex transformations between $\mathbf{v}_s$ and $\hat{\mathbf{x}}_s$. However, the introduction of learnable $g_{RAM}$ requires additional computations/iterations for parameter adjustment \cite{Beal_2022_WACV}, when the optimization space of $\theta$ is larger than the optimization space of $\mathbf{v}_s$, which is often the case. Moreover, the capability for powerful nonlinear transformations brought about by a large number of learnable parameters may cause generation space collapse, leading to all $V_s$ being transformed to the same $\hat{\mathbf{x}}_s$. In this case, additional efforts are needed to maintain the internal structure of $\hat{X}_s$ \cite{zbontarBarlowTwinsSelfSupervised,zhuTiCoTransformationInvariance2022}.

\subsection{Dynamic Optimization Scheduling}
\label{dynamic scheduel}
The last puzzle in EEE is the dynamic scheduling of the optimization process. In general, this refers to the group of hyperparameters that are responsible for controlling the pace of cooperation between the various modules, including the early stop threshold $\epsilon_{es}$, the RAM net iteration round $r_{RAM}$, and the focus coefficient $c_f$.

In the proposed framework, there is a requirement that $g_e$ undergoes much more training compared to $g_{RAM}$. This is because the training of $g_e$ is on the incremental set $B_e \cup B_h$, and its good performance is a prerequisite for the successful operation of $g_{RAM}$. Note that the validation process is considered time-consuming, we suppose the time for one query of $f_v$ can afford $r_t$ training rounds of  $g_e$. Meanwhile, $r_b$ training rounds for $g_e$ are expected to suffice in order to reflect the distribution of $B_e \cup B_h$. Therefore, without considering the time needed for exploring and searching a state, there is $r_{ep}=\frac{r_t}{r_b}$ training epochs. However, excessive training of $g_e$ may lead to overfitting, which should be avoided. Early stopping, a simple yet useful tip, is used to prohibit this. Although using hold out for early stopping \cite{camorianoNYTROWhenSubsampling} is available, it may lead to loss of very limited data, and thus, we choose to terminate the training of a $g_e$ and regard its training as convergent when the average loss in one epoch is smaller than a predefined $\epsilon_{es}$.

The number of convergent error estimators $l, l\le L$ can represent the confidence of the entire $G_e$. Therefore, we can correspondingly set $r_{RAM}\propto l$. Because the more confident $G_e$ is, the more confidently we can iterate $g_{RAM}$. Indeed, the time savings through early stopping of the training of $g_e$, also allows for additional training rounds to iterate $g_{RAM}$. 

The focus coefficient $c_f, c_f>1$ is the switch to control whether the searched state $\hat{\mathbf{x}}_c$ should be passed to query the validation module $f_v$. Because $f_v$ is computationally expensive, it is not efficient to consider each $\hat{\mathbf{x}}_c$. Instead, we only query $\hat{\mathbf{x}}_c$, when $\hat{e}_o(\hat{\mathbf{x}}_c)<c_f*\epsilon$. One can set $c_f \propto \frac{1}{l}$, since the smaller the $l$ is, the higher the degree of uncertainty of $\hat{e}_o(\hat{\mathbf{x}}_c)$, so that querying the validation module for calibration is more needed. Of course, one can also set $c_f$ to be constant for convenience.

\section{Experiments}
\subsection{Task Description}
We tested EEE on four real-world engineering problems. These are gas quantification in Laser Absorption Spectroscopy (LAS), turbofan engine design, electro-mechanical actuator design, and pulse-width modulation of 13-level inverters. These problems were chosen as they have diverse state dimensions $D_x$, which are respectively equal to 2, 11, 20, and 30. Thus, a comprehensive evaluation of the properties and performance of EEE can be given. 

\textbf{Problem 1: Gas quantification in LAS} was taken from \cite{kangSelfValidatedPhysicsEmbeddingNetwork2022}. The problem involves the estimation of the temperature $T_m$ and concentration $C_m$ of the molecule of interest by using LAS measurements $S_m$. In \cite{kangSelfValidatedPhysicsEmbeddingNetwork2022}, the molecule is CO$_2$, and the spectrum dimension is 200. The task for a pretrained model $f_p$ is under a given $\mathbf{y}_0: S_m \in \mathbb{R}^{200}$, to estimate $\mathbf{x}_0: [T_m, C_m]\in \mathbb{R}^2$ \cite{renMachineLearningApplied2019}.

\textbf{Problem 2: Turbofan Engine design} was also taken from \cite{kangSelfValidatedPhysicsEmbeddingNetwork2022}. It is a design problem, which involves the determination of a group of cycle and component parameters $\mathbf{x}_{cac}$ to satisfy the engine performance requirement, i.e., thrust force $F_t$ and thrust specific fuel consumption $C_{tf}$. Therefore, for a pretrained model $f_p$, the task is according to the performance requirement  $\mathbf{y}_0: [F_t, C_{tf}] \in \mathbb{R}^2$ to produce a group of cycle and component parameters $\mathbf{x}_0:\mathbf{x}_{cac} \in \mathbb{R}^{11}$ \cite{degiorgiJetEngineDegradation2019}.

\textbf{Problem 3: Electro-mechanical actuator design} We considered the "CS1" problem from \cite{picardRealisticConstrainedMultiobjective2021}. This requires the design of an electro-mechanical actuator with 20 design parameters $\mathbf{x}_d \in \mathbb{R}^{20}$ to maximize the safety factor requirement $C_s$ and  minimize the total cost $C_t$ under seven inequality constraints $\mathbf{e}_{ie} \in \mathbb{R}^7$. Therefore, for a potential pretrained model $f_p$, the problem can be set as follows: according to different safety factor and total cost requirements $\mathbf{y}_0: [C_s,C_t] \in \mathbb{R}^2$,  find a group of suitable design parameters $\mathbf{x}_0:\mathbf{x}_d \in \mathbb{R}^{20}$.

\textbf{Problem 4: Pulse-width Modulation of 13-level Inverters} Here, we consider the "RCM35" problem, collected by \cite{kumarBenchmarkSuiteRealWorldConstrained2021} from \cite{edpugantiOptimalPulsewidthModulation2017}. The problem is to manipulate 30 control parameters  $\mathbf{x}_{cp} \in \mathbb{R}^{30}$ to minimize the distortion factor $F_d$ and nonlinear constraint $C_n$. Moreover, 29 $\mathbf{e}_{ie}$ exist, which can be expressed as  $\mathbf{x}_{cp,i}-\mathbf{x}_{cp,i+1} \le 0$. A model $f_p$ can be pretrained to have the function that, according to different levels of emphasis on  the distortion factor $F_d$ and nonlinear constraint $nc$, i.e., $\mathbf{y}_0: [F_d,C_n] \in \mathbb{R}^2$, to find the best control plan $\mathbf{x}_0:\mathbf{x}_{cp} \in \mathbb{R}^{30}$ \cite{durgasukumarComparisonAdaptiveNeuroFuzzybased2012}.

In our experiments, we suppose that all pretrained $f_p$s have already failed, and EEE is used to find the validation-viable $\hat{\mathbf{x}}$. As for problems 1 and 2, we directly utilized the test cases provided by \cite{kangSelfValidatedPhysicsEmbeddingNetwork2022}. For problem 1, the test cases contain 200 spectra $Y_0$ generated within the range of $[\mathbf{x}_0: 700 K<t_m<2000 K, 0.01<c_m<0.5]$, and the feasible range of $\mathbf{x}$ for each case was also predefined. For problem 2, the test case is to estimate a group of $\mathbf{x}_{cac}$ to realize the performance of CFM-56 \cite{kumarInnovativeApproachesReducing2012}. In this respect, we applied optimization algorithms, including EEE, to run 100 tests with different initial random sampling, respectively.  Problems 3 and 4 are originally designed for the testing of evolutionary algorithms. We randomly sampled 100 samples from the known Pareto fronts of each problem, respectively, to demonstrate the varying degrees of emphasis on the two objectives in each problem. All elements of $\mathbf{x}$ were treated as continuous rather than discrete variables.

\subsection{EEE Configuration}
\label{configuration}
\textbf{Validation module construction} All of the problems considered in this work have available forward models $f_f:\mathcal{X} \to \mathcal{Y}$. Thus, the first validation process is set as $e_d=d(\hat{\mathbf{y}}_0,\mathbf{y}_0)$, for problem 1. This is done on the entire spectrum, as all intensities are correlated, and $d$ is chosen to be the L2 loss. However, for problems 2,3, and 4, $e_d$ is calculated on each entry of $\hat{\mathbf{y}}_0$, as they are independent, and $d$ is chosen to be the L1 loss. Note that we consider the simulation process is time-consuming. We choose these four computationally efficient problems merely for the convenience of experiments.
For problem 2, we used the technology mismatch error from \cite{kangSelfValidatedPhysicsEmbeddingNetwork2022}: $e_e= [\sigma (\hat{\mathbf{x}}_{cac})]$. As for problems 3 and 4, we also used the mentioned inequality constraints \cite{picardRealisticConstrainedMultiobjective2021,kumarBenchmarkSuiteRealWorldConstrained2021} as part of $f_v$. Moreover, after normalizing the feasible domain of $\mathbf{x}$ to be [0,1], we formulated the validation process of the boundary error $\mathbf{e}_b=max(\hat{\mathbf{x}}_{norm}-1,0)+max(-\hat{\mathbf{x}}_{norm},0)$, so that $g_{RAM}$ can learn the knowledge of the feasible domain through punishment. A summary of the validation module is shown in Table \ref{table 1}, for problems 1 and 2. The $e_o$ here is consistent with that of\cite{kangSelfValidatedPhysicsEmbeddingNetwork2022}. In problem 3, the constraints are treated as soft ones, while for problem 4, they are treated as rigid ones with large weights. Indeed, it is possible to change the formulation style of the constraints according to the characteristics of the problem. 

\begin{table*}[]
\centering
\caption{Summary of validation module configuration}
\label{table 1}
\resizebox{\textwidth}{!}{%
\begin{tabular}{ccccc}
\hline
\textbf{Problem} & \textbf{1} & \textbf{2} & \textbf{3} & \textbf{4} \\ \hline
$D_x$ & 2 & 11 & 20 & 30 \\
$\mathbf{e}_{imp}$ & $e_d \in \mathbb{R}$ & $\mathbf{e}_d\in \mathbb{R}^2$ & $\mathbf{e}_d\in \mathbb{R}^2$; $\mathbf{e}_{ie}\in \mathbb{R}^7$ & $\mathbf{e}_d\in \mathbb{R}^2$ \\
$\mathbf{e}_{exp}$ & $\mathbf{e}_b \in \mathbb{R}^2 $ & $\mathbf{e}_b \in \mathbb{R}^{11} $;$e_{e} \in \mathbb{R} $ & $\mathbf{e}_b \in \mathbb{R}^{20} $ & $\mathbf{e}_b \in \mathbb{R}^{30} $; $\mathbf{e}_{ie}\in \mathbb{R}^{29}$ \\
$D_e$ & 3 & 14 & 29 & 61 \\
$e_o$ & $e_d+\mathbb{E}(\mathbf{e}_b)$ & $\mathbb{E}(\mathbf{e}_d)+0.1e_e+0.1\mathbb{E}(\mathbf{e}_b)$ & $\mathbb{E}([\mathbf{e}_d,\mathbf{e}_{ie}])+0.1\mathbb{E}(\mathbf{e}_b)$ & $0.5\mathbb{E}(\mathbf{e}_d)+10\mathbb{E}(\mathbf{e}_{ie})++0.1\mathbb{E}(\mathbf{e}_b)$ \\
$\epsilon$ & 0.1 & 0.05 & 0.05 & 0.05 \\ \hline
\end{tabular}%
}
\end{table*}

\textbf{Networks and hyperparameters} Both $g_e$ and $g_{RAM}$ used in all four problems were almost the same. In the case of $g_e$, this was chosen to be a simple MLP with the architecture of $D_x \to h \to 2h \to h \to D_{e_{imp}}$. Here, $h$ refers to the number of hidden neuron, and set to 1024 for all problems. ReLu was chosen as the activation function in the hidden layers, while sigmoid functions were used in the output layers, as the error is non-negative. By default, four $g_e$s were utilized, to save on the use of computational resources. A larger number of modules can be considered if training becomes parallelized, thus alleviating computational times. 

Considering the kernel view of $g_{RAM}$, three cases were considered: (1) Identity kernel, where the optimization is directly performed on  $\mathbf{v}_s \leftrightarrow \hat{x}_s$, (2) Perceptron kernel, where a single layer network $p: \mathbb{R} \to \mathcal{X}$ was used, (3) MLP kernel, where the structure $g_e$ is directly inherited, however, both input and output sizes were set to $D_x$. The number of hidden units $h$ was set to 20 for problems 1, 2 and 3, and 256 for problem 4 in order to illustrate generation space collapse.  Problem 4, in fact, can only search states inside a hyper-triangle dominated by the first entry of state $\hat{\mathbf{x}}_1$. Therefore, we defined a special function to reflect this reality by processing the output of $g_{RAM}$ :$\hat{\mathbf{x}}_{norm,i}=\hat{\mathbf{x}}_{norm,1}+\frac{1}{1+exp(-\sum^i_1\hat{\mathbf{x}}_{norm,i})}(1-\hat{\mathbf{x}}_{norm,1})$. This function was embedded directly into the MLP- and perceptron-type $g_{RAM}$, and also utilized for the identity-kernel-type $g_{RAM}$ and other optimization algorithms introduced later. Therefore, the error transfer uncertainty was further reduced. To avoid generation collapse, a simple regularization term was added to $\mathcal{L}_{RAM}$ in problem 4. This is defined as $e_{gc}=max(\frac{0.288}{10}-\sigma (\hat{\mathbf{x}}_{norm,1}),0)$, which reflects the requirement that the generated states should not collapse to an identical state, where 0.288 is the $\sigma$ of the uniform distribution. 

In the case of CEE, a perceptron $f_{\theta}: \mathbb{R} \to \mathcal{X}$ was used to simulate a hyper-line. One epoch contained 40 batches of training, with one batch containing 64 samples. In cases where the size of the initial sampling was smaller than 64, the batch size was gradually increased to 64. Parameter $\epsilon_{es}=1e^{-3}$ by default, $r_{RAM}=[int(2l/L)+1]*r_b$, where $r_b=1$ for identity-kernel- and perceptron-type $g_{RAM}$, as there is a small number of learnable parameters, however, $r_b=10$ for MLP-type $g_{RAM}$. The focus coefficient $c_f$ was set to be 1.5, 1.5, 2, and 5 for problems 1, 2, 3 and 4, respectively, as they are of increasing degrees of complexity, and thus, the confidence on estimation accuracy of $g_e$ was correspondingly decreased. In summary, we did not attempt to carefully optimize  the utilized networks and hyperparameters. Indeed, there were very limited differences between the models used in the four problems. This is because the characteristics of the given optimization problems cannot be thoroughly recognized prior to processing, which provides very limited prior information to adjust the hyperparameters. Therefore, the proposed framework needs to be robust in the case of unseen problems with limited prior knowledge.  

\subsection{Main Results} 
We compared EEE with SVPEN\cite{kangSelfValidatedPhysicsEmbeddingNetwork2022}, and several traditional and advanced optimization methods, i.e, Bayesian Optimization with Gaussian Process (GP), GA\cite{mirjaliliGeneticAlgorithm2019}, PSO \cite{kennedy1995particle}, CMAES \cite{hansenCMAEvolutionStrategy2006}, ISRES \cite{runarssonSearchBiasesConstrained2005}, NSGA2 \cite{debFastElitistMultiobjective2002}, and UNSGA3 \cite{seadaUnifiedEvolutionaryOptimization2016}. We kept the default setting of SVPEN in problems 1 and 2, and constructed SVPEN based on the $g_{RAM}$ and $g_e$ in problems 3 and 4. The realizations of other optimizations were based on Bayesian Optimization \cite{bayesian} and Pymoo \cite{blank2020pymoo}. The efficiency and convergence of these methods were evaluated. The efficiency was measured by the metrics of average query times of the validation module $\Bar{t}$, and median query times $t_{0.5}$. The convergence was mainly measured by the metrics of the standard deviation of query times $\sigma_t$, and the failed runs in all test cases $r_{f}$. In each test case, the algorithm was allowed to query the validation module up to 1000 times. For example, for GP, validation and exploration were combined into a single query, while EEE needed two individual queries. We demonstrated conditions starting with 32 or 64 initial samples by random sampling. The population of algorithms in GA was correspondingly set with this initial size. The metrics were calculated without considering these initial samples though CMAES were incompatible with the initial sampling. 
  
 The results are shown in Table \ref{table 2}, where bold numbers indicate the best, bold-italic numbers indicate the sub-best, and "-" indicates thorough failure. Only the best of the three types of $g_{RAM}$ is presented here. For problems 1,2, and 3, these are the identity kernels; however, for problem 4, it is the MLP. Under the condition of 64 initial samples, EEE has sub-best efficiency in problems 1,2, and 3, and best $t_{0.5}$ in problem 4. Although it does not have a $\sigma_t$ as small as others, the smaller $\sigma_t$ in other algorithms except for GP is based on a much larger number of queries. Notably, EEE has good applicability in all four diverse problems, while other algorithms are clearly not suited to some problems; for example, GP thoroughly failed in problem 3. More importantly, EEE always has the least $r_f$, which is the intrinsic property of the complementary system we want. It is notable that SVPEN, another network-based algorithm, can only work decently in problem 1, although in most runs of other problems, the error has a slowly descending trend. This is to be expected as its design does not consider the requirements of efficiency and convergence. Although the performance of EEE degrades significantly, when the initial sampling drops to 32, it still demonstrates competitive performance in either convergence or efficiency in all four problems.

 In fact, EEE shares some similarities to GP and genetic algorithms. From one perspective, we tried to make the error transfer as precise and efficient as GP. However, the nature of the network cannot do so perfectly; thus, GP is more appealing for simple low-dimensional optimization problems, while with an intelligent search model, EEE can demonstrate better performance in high-dimensional optimization tasks. From another perspective, the ensemble search of $g_{RAM}$ is a kind of gradient-based genetic algorithm, but the validation can be practised firstly in $G_e$ to select the most prosperous state. However, ordinary genetic algorithms have to query $f_v$ frequently, and although they may only need one round of queries, the relatively large population will still result to heavy computational expenses. 

\begin{table*}[]
\caption{Comparison between EEE and state-of-the-art optimization methods}
\label{table 2}
\resizebox{\textwidth}{!}{%
\begin{tabular}{|cc|cccc|cccc|cccc|cccc|}
\hline
\multirow{2}{*}{\textbf{\begin{tabular}[c]{@{}c@{}}Initial\\     samples\end{tabular}}} & \multirow{2}{*}{\textbf{Algorithm}} & \multicolumn{4}{c|}{\textbf{Problem 1}} & \multicolumn{4}{c|}{\textbf{Problem 2}} & \multicolumn{4}{c|}{\textbf{Problem 3}} & \multicolumn{4}{c|}{\textbf{Problem 4}} \\
 &  & $t_{0.5}$ & $\Bar{t}$ & $\sigma_t$ & $r_f$ & $t_{0.5}$ & $\Bar{t}$ & $\sigma_t$ & $r_f$ & $t_{0.5}$ & $\Bar{t}$ & $\sigma_t$ & $r_f$ & $t_{0.5}$ & $\Bar{t}$ & $\sigma_t$ & $r_f$ \\ \hline
\multirow{9}{*}{64} & \textbf{EEE(ours)} & \textit{\textbf{5}} & \textit{\textbf{4.89}} & 2.48 & 0 & \textit{\textbf{13}} & \textit{\textbf{20.20}} & 16.37 & 0 & \textit{\textbf{148.5}} & 189.90 & 164.96 & \textbf{0} & \textit{\textbf{46.5}} & \textbf{81.26} & \textbf{155.30} & \textbf{2} \\
 & GP & \textbf{1} & \textbf{1.40} & \textit{\textbf{0.69}} & 0 & \textbf{10} & \textbf{9.24} & \textit{\textbf{3.97}} & 0 & - & - & - & - & \textbf{33} & \textit{\textbf{227.63}} & 364.08 & \textit{\textbf{16}} \\
 & GA & 64 & 64.32 & 4.51 & 0 & 64 & 64.00 & \textbf{0} & 0 & 352 & 353.28 & \textit{\textbf{105.74}} & \textbf{0} & 128 & 297.92 & 363.45 & 20 \\
 & PSO & 64 & 64.00 & \textbf{0} & 0 & 64 & 64.00 & \textbf{0} & 0 & \textbf{128} & \textbf{157.84} & 137.40 & \textit{\textbf{1}} & 64 & 290.96 & 373.65 & 18 \\
 & CMAES & 19 & 24.61 & 19.28 & 0 & 78 & 77.56 & 4.38 & 0 & 271 & 302.59 & 156.24 & \textit{\textbf{1}} & 197 & 344.54 & \textit{\textbf{363.18}} & 22 \\
 & ISRES & 65 & 65.96 & 7.78 & 0 & 193 & 193.00 & \textbf{0} & 0 & 385 & 391.54 & 241.22 & 3 & 129 & 313.69 & 368.78 & 19 \\
 & NSGA2 & 64 & 64.32 & 4.51 & 0 & 64 & 64.00 & \textbf{0} & 0 & 320 & 352.00 & 114.31 & \textbf{0} & 128 & 299.84 & 364.63 & 20 \\
 & UNSGA3 & 64 & 64.32 & 4.51 & 0 & 64 & 64.00 & \textbf{0} & 0 & 384 & 368.64 & \textbf{102.85} & \textbf{0} & 128 & 310.72 & 370.24 & 20 \\
 & SVPEN & 74 & 330.45 & 409.89 & 74 & - & - & - & - & - & - & - & - & - & - & - & - \\ \hline
\multirow{8}{*}{32} & \textbf{EEE(ours)} & \textit{\textbf{5}} & \textit{\textbf{5.47}} & \textit{\textbf{4.39}} & 0 & \textit{\textbf{31}} & 33.63 & 19.35 & 0 & 197.5 & \textit{\textbf{233.96}} & 180.01 & \textbf{0} & 64 & \textbf{167.77} & \textbf{284.31} & \textbf{10} \\
 & GP & \textbf{1} & \textbf{1.71} & \textbf{1.02} & 0 & \textbf{11} & \textbf{9.60} & 3.89 & 0 & - & - & - & - & \textit{\textbf{42}} & \textit{\textbf{239.12}} & 367.12 & 15 \\
 & GA & 32 & 33.60 & 8.31 & 0 & 32 & \textit{\textbf{32.00}} & 0.00 & 0 & 224 & 241.60 & \textbf{71.75} & \textbf{0} & 224 & 241.60 & 382.51 & 21 \\
 & PSO & 32 & 32.32 & 4.51 & 0 & 32 & \textit{\textbf{32.00}} & 0.00 & 0 & \textbf{160} & 311.20 & 333.45 & 18 & 96 & 270.80 & 324.54 & \textit{\textbf{14}} \\
 & CMAES & 19 & 24.61 & 19.28 & 0 & 78 & 77.56 & 4.38 & 0 & 271 & 302.59 & 156.24 & \textit{\textbf{1}} & 160 & 283.28 & \textit{\textbf{363.18}} & 22 \\
 & ISRES & 32 & 36.00 & 13.93 & 0 & 64 & 64.00 & 0.00 & 0 & 352 & 416.24 & 209.23 & 3 & 141 & 320.83 & 386.75 & 21 \\
 & NSGA2 & 32 & 33.92 & 10.44 & 0 & 32 & \textit{\textbf{32.00}} & 0.00 & 0 & 224 & 239.44 & 150.26 & \textit{\textbf{1}} & 64 & 276.24 & 394.99 & 22 \\
 & UNSGA3 & 32 & 34.08 & 13.98 & 0 & 32 & \textit{\textbf{32.00}} & 0.00 & 0 & \textit{\textbf{192}} & \textbf{218.72} & \textit{\textbf{136.53}} & 2 & \textbf{32} & 262.88 & 396.51 & 22 \\ \hline
\end{tabular}%
}
\end{table*}

\subsection{Ablation Studies}
To avoid the effect of auxiliary randomness, we performed ablation studies focusing mainly on problem 2. Without further explanation, the baseline model here is the original identity-kernel-style $g_{RAM}$, but with merely two error estimators. 

\textbf{Error informativity and uncertainty} We compared the baseline model with two conditions;(a) Using the overall error, i.e., $e_o$ was used to guide the updates, and (b) All error components were simulated. Network $g_e$ was used to simulate the errors. The results in Table \ref{table 3} demonstrate that low-informativity error (i.e., scalar $e_o$) degrades optimization efficiency. However, without reducing error transfer uncertainty, making $G_e$ to transfer all the elements of $\mathbf{e}$ is even more harmful. This demonstrates the importance of coordination between error informativity and error uncertainty.

\begin{table}[!htbp]
\centering
\caption{Problem 2: effect of error informativity and uncertainty}
\label{table 3}
\begin{tabular}{llll}
\hline
\multicolumn{1}{c}{\textbf{Condition}} & \textbf{$t_{0.5}$} & \textbf{$\Bar{t}$} & \textbf{$\sigma_t$} \\ \hline
Baseline & \textbf{13.00} & \textbf{20.20} & \textbf{16.37} \\
Overall error & 16.00 & 23.26 & 21.18 \\
Simulate all errors & 55.50 & 62.22 & 57.08 \\ \hline
\end{tabular}
\end{table}

\textbf{Ensemble of error estimators} The number of error estimators $L$ was increased firstly to four, and next to eight, however, only two $g_e$s were used for state exploration. The results in  Table \ref{table 4}  show that with the increase in $L$, both convergence and efficiency are improved, but the improvement in efficiency becomes saturated, after increasing the number of error estimators to some level.

\begin{table}[!htbp]
\centering
\caption{Problem 2: effect of the number of error estimators}
\label{table 4}
\begin{tabular}{cccc}
\hline
\textbf{$g_e$ number} & \textbf{$t_{0.5}$} & \textbf{$\Bar{t}$} & \textbf{$\sigma_t$} \\ \hline
Two(Baseline) & 13.00 & 20.20 & 16.37 \\
Four & \textbf{7.00} & \textbf{15.44} & 13.86 \\
Eight & 8.00 & 15.09 & \textbf{13.01} \\ \hline
\end{tabular}
\end{table}

\textbf{Exploration method} CEE was compared to model-based active exploration \cite{pathakSelfSupervisedExplorationDisagreement} with two, four and eight MLP-type $g_e$s, respectively. The metric of the maximum number of queries to the validation module $t_{max}$ was used to demonstrate the difference in optimization convergence. Table \ref{table 5} presents the results, where "active-2" represents active model-based exploration with two $g_e$. The results of $t_{max}$ and $\sigma_t$  demonstrate that model-based active learning makes optimization prone to become unstable. Although model-based active exploration provides an advantage in $t_{0.5}$, the disadvantages in convergence make it perform horribly in $\Bar{t}$.

\begin{table}[!htbp]
\centering
\caption{Problem 2: effect of exploration method}
\label{table 5}
\begin{tabular}{ccccc}
\hline
\textbf{Search} & \textbf{$t_{0.5}$} & \textbf{$\Bar{t}$} &\textbf{$t_{max}$} & \textbf{$\sigma_t$}\\ \hline
\begin{tabular}[c]{@{}c@{}}CEE-2\end{tabular} & 30.50 & 31.87 & 115.00 & \textbf{22.75} \\
CEE-4 & 28.00 & 32.64 & \textbf{112.00} & 22.82 \\
CEE-8 & 26.00 & \textbf{31.11} & 210.00 & 29.82 \\
Active-2 & 23.00 & 45.08 & 968.00 & 101.65 \\
Active-4 & 24.00 & 41.32 & 959.00 & 97.15 \\
Active-8 & \textbf{20.00} & 31.22 & 509.00 & 52.25 \\ \hline
\end{tabular}
\end{table}

\textbf{Ensemble search} We checked the effect of ensemble state search by changing the number of seed states. The results shown in Table \ref{table 6} show that utilizing an ensemble of seed states significantly outperforms using a single seed state in terms of both efficiency and convergence. However, too many seeds may harm efficiency, although convergence is still improved. One potential explanation is that more seed states present a higher probability to generate adversarial states, which get low $\hat{e}_o$ from $G_e$ but instead acquire a much higher $e_o$ due to inconsistencies between $G_e$ and $f_v$. The presence of this condition may disturb optimization efficiency. However, a larger number of seeds can always reduce the partition of states falling into local optima or divergence, so convergence is improved.

\begin{table}[!htbp]
\centering
\caption{Problem 2: effect of ensemble search }
\label{table 6}
\begin{tabular}{ccccc}
\hline
\textbf{Seed number} & \textbf{$t_{0.5}$} & \textbf{$\Bar{t}$} &\textbf{$t_{max}$} & \textbf{$\sigma_t$} \\ \hline
1 & 69.00 & 72.56 & 195.00 & 36.13 \\
32 & 24.00 & 28.21 & 87.00 & 17.05 \\
64 & \textbf{13.00} & \textbf{20.20} & 86.00 & 16.37 \\
128 & 25.00 & 26.95 & \textbf{80.00} & \textbf{14.12} \\ \hline
\end{tabular}
\end{table}

\textbf{Search model type and generation space collapse} We compared the performance of all three types of $g_{RAM}$, as shown in Table \ref{table 7}. This shows that identity-kernel-type $g_{RAM}$ outperforms the other two types in problem 2. This is because this problem is relatively simple, so a learnable system merely increases optimization expense. Perceptron performs worst as it has even fewer changeable parameters (eleven, in fact) than the identity-kernel-type model (22), and its optimization is basically done on a rotating and elastic hyper-line. 

However, for problem 4, without using the state space regularization term mentioned in Sec.\ref{configuration}, the MLP works better than the identity-kernel-type $g_{RAM}$. This is because $D_x$ is relatively high and correlations exist between the entries of $\mathbf{x}$, so a rather large model is preferred to tackle complex transformations. On the other hand, with a relatively large MLP, there is a higher likelihood for the occurrence of generation space collapse, evidenced by the poor convergence of the MLP. By simply adding the state space regularization term, convergence improves, whereas $t_{0.5}$ is correspondingly weakened, as this regularization term affects the efficiency requirement.

\begin{table}[!htbp]
\centering
\caption{Effect of search model type and generation space collaspe}
\label{table 7}
\begin{tabular}{cccccc}
\hline
\textbf{Problem} & \textbf{$g_{RAM}$} &\textbf{$t_{0.5}$} & \textbf{$\Bar{t}$} &\textbf{$t_{max}$} & \textbf{$\sigma_t$} \\ \hline
\multirow{3}{*}{2} & identical kernel & \textbf{13.00} & \textbf{20.20} & \textbf{16.37} & 0 \\
 & MLP & 30.50 & 31.87 & 22.75 & 0 \\
 & perception & 34.00 & 47.82 & 42.55 & 0 \\ \hline
\multirow{3}{*}{4} & identical kernel & 94.50 & 206.93 & 307.01 & 11 \\
 & MLP w/o reg & \textbf{46.00} & 128.13 & 262.59 & 8 \\
 & MLP-reg & 55.50 & \textbf{88.17} & \textbf{165.82} & \textbf{3} \\ \hline
\end{tabular}
\end{table}

\textbf{Dynamic schedule} We compared the baseline case with the other four variants, namely, (a) $\epsilon_{es}=1e^{-3}$, (b) $\epsilon_{es}=1e^{-5}$, (c) $r_{RAM}=3$, the maximum value of original schedule, and (d) removing $c_f$. The results, summarized in Table \ref{table 8}, demonstrate that either decreasing or increasing $\epsilon_{es}$ is harmful in terms of efficiency, although EEE still works, which embodies the tradeoff between bias and variance is pretty subtle. In addition, too many runs of $g_{RAM}$ harm the efficiency of optimization, because the biased estimation of $G_e$ may mislead the optimization direction of $g_{RAM}$. Without using $c_f$, the direct result is that many redundant queries of the validation module are carried out, thus, slowing down the optimization efficiency.

\begin{table}[!htbp]
\centering
\caption{Problem 2: effect of dynamic schedule }
\label{table 8}
\begin{tabular}{cccc}
\hline
\textbf{schedule} & \textbf{$t_{0.5}$} & \textbf{$\Bar{t}$} & \textbf{$\sigma_t$} \\ \hline
baseline & \textbf{13} & \textbf{20.20} & 16.37 \\
$\epsilon_{es}=1e^{-3}$& 38 & 37.55 & 17.28 \\
$\epsilon_{es}=1e^{-5}$& 23 & 26.20 & \textbf{15.45} \\
$r_{RAM}=3$ & 30 & 32.80 & 17.84 \\
w/o $c_f$ & 23 & 27.19 & 19.36 \\ \hline
\end{tabular}
\end{table}

\section{Discussion and Conclusion}
In this work, EEE, a network-based optimization method to remediate the failure of pretrained machine learning models was proposed. The problem tackled herein required an online optimization method to provide a validation-viable state estimate with the least possible number of queries to the validation module and the highest possible success rate. To satisfy the requirements of efficiency and convergence, the key is the error function from the validation module. In EEE, we focused our endeavors mainly around three aspects of error: (1) Error ambiguity was reduced by constructing high dimensional error functions to increase error informativity, (2) Error transfer uncertainty was reduced by categorizing error components into implicit and explicit. Differentiable expressions were used to directly calculate explicit errors and an ensemble of error estimators $G_e$ was used to reduce the approximation uncertainty of implicit errors. Moreover, Constrained Ensemble Exploration (CEE) was introduced to accelerate the learning of $G_e$ and stabilize the optimization process. (3) Effective error utilization was enabled, through a greedy-search-assisted ensemble state search model $g_{RAM}$, which can efficiently search for the most prosperous states. From the viewpoint of the kernel, we discussed the characteristics of diverse search model types and the means for reducing the occurrence of state generation space collapse. 

Although EEE is as competitive to or outperforms current popular optimization algorithms in four real-world engineering problems, there is room for improvement. Due to the nature of neural networks, parameter tuning requires time and data. The current success of EEE is based on the hypothesis that the validation module, in science and engineering applications, often encapsulates computationally expensive simulation processes. In contrast, the training expenses of EEE are negligible. Currently, the optimization process of EEE is very efficient, as the number of learnable parameters in both $g_e$ and $g_{RAM}$ is reasonable. However, in extremely complex problems with very high state dimensionality, a larger model seems inevitable, and optimization costs would be of concern. In these cases, the effectiveness of state generation from both $g_{RAM}$ and CEE may also be a problem, because the effective samples are sparsely distributed in high-dimensional state space. The interesting thing is that the flexibility of network architecture allows for embedding domain/induced knowledge and complex operations to reflect the characteristics of the problem. This, in turn, makes neural networks the method of choice in solving high-dimensional optimization problems.

%Bibliography
\bibliographystyle{unsrt}  
\bibliography{references}

\end{document}